\documentclass[letterpaper]{article} 
\usepackage{aaai25}  
\usepackage{times}  
\usepackage{helvet}  
\usepackage{courier}  
\usepackage[hyphens]{url}  
\usepackage{graphicx} 
\usepackage{natbib}  
\usepackage{caption} 
\usepackage{algorithm}
\usepackage{algorithmic}

\usepackage{newfloat}
\usepackage{listings}
\DeclareCaptionStyle{ruled}{labelfont=normalfont,labelsep=colon,strut=off} 
\floatstyle{ruled}
\newfloat{listing}{tb}{lst}{}
\floatname{listing}{Listing}
\usepackage{xcolor}    
\definecolor{blue}{rgb}{0,0,1}
\usepackage{amsmath} 
\usepackage{amssymb}
\usepackage{booktabs} 

\title{GLAM: Global-Local Variation Awareness in \\
Mamba-based World Model}
\author{
    Qian He\textsuperscript{1,2,3}\equalcontrib, Wenqi Liang\textsuperscript{1,2,3}\equalcontrib, Chunhui Hao\textsuperscript{4}, Gan Sun\textsuperscript{5}, Jiandong Tian\textsuperscript{1}\thanks{Corresponding author}\\
}
\affiliations{
    \textsuperscript{1}State Key Laboratory of Robotics, Shenyang Institute of Automation, Chinese Academy of Sciences.\\
    \textsuperscript{2}Institutes for Robotics and Intelligent Manufacturing, Chinese Academy of Sciences.\\
    \textsuperscript{3}University of Chinese Academy of Sciences. \\
    \textsuperscript{4}Shenyang University of Chemical Technology.\\
    \textsuperscript{5}School of Automation Science and Engineering, South China University of Technology.
    
    \{heqian0325, liangwenqi0123, sungan1412\}@gmail.com, \{haochunhui, tianjd\}@sia.cn
}

\usepackage{bibentry}

\begin{document}

\maketitle

\begin{abstract}
Mimicking the real interaction trajectory in the inference of the world model has been shown to improve the sample efficiency of model-based reinforcement learning (MBRL) algorithms. 
Many methods directly use known state sequences for reasoning. However, this approach fails to enhance the quality of reasoning by capturing the subtle variation between states. Much like how humans infer trends in event development from this variation, in this work, we introduce \textbf{G}lobal-\textbf{L}ocal variation \textbf{A}wareness \textbf{M}amba-based world model (GLAM) that improves reasoning quality by perceiving and predicting variation between states. GLAM comprises two Mamba-based parallel reasoning modules, GMamba and LMamba, which focus on perceiving variation from global and local perspectives, respectively, during the reasoning process. GMamba focuses on identifying patterns of variation between states in the input sequence and leverages these patterns to enhance the prediction of future state variation. LMamba emphasizes reasoning about unknown information, such as rewards, termination signals, and visual representations, by perceiving variation in adjacent states. By integrating the strengths of the two modules, GLAM accounts for higher-value variation in environmental changes, providing the agent with more efficient imagination-based training. We demonstrate that our method outperforms existing methods in normalized human scores on the Atari 100k benchmark.

\end{abstract}
\begin{links}
    \link{Code}{https://github.com/GLAM2025/glam}
\end{links}

\section{Introduction}

\begin{figure}[ht]
\centering
\includegraphics[width=234pt,]
{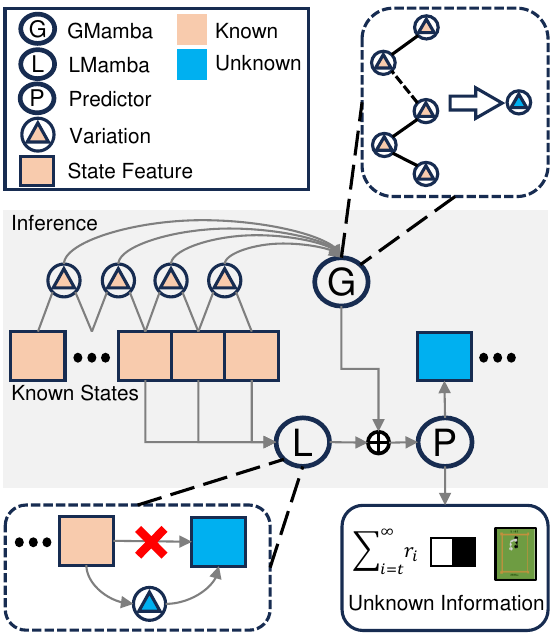}
\vspace{-6pt}
\caption{A single-step demonstration of GLAM inference. GLAM leverages GMamba and LMamba for global and local state variation, respectively. GMamba captures patterns of variation changes within the sequence, while LMamba incorporates variation awareness when inferring unknown information, rather than relying on direct inference alone. } 
\vspace{-0.15in}
\label{fig: small}
\end{figure}

A key challenge in deep reinforcement learning (DRL) is the lack of sample efficiency \cite{schrittwieser2020mastering,vinyals2019grandmaster}. Many model-based reinforcement learning (MBRL) \cite{sutton1991dyna} algorithms tackle this challenge by learning models of the environment to improve the training efficiency of agents. Inspired by human cognitive processes, an MBRL approach trains a network known as a \textit{world model} to simulate the environment and subsequently train agents using this model. The concept of world model has achieved breakthroughs in various decision-making domains.

These algorithms utilize interaction data between the agent and the real environment to train a parameterized world model. The agent can generate an unlimited amount of training data through interactions with this world model, thus enhancing its decision-making capabilities. This method of generating data is analogous to \textit{imagination} of the human. However, this imagination-based self-supervised learning often fails if the data provided by the world model are not sufficiently accurate. 
Therefore, the lack of accuracy in imagination is a bottleneck for the world model to improve sample efficiency.

Numerous endeavors have been made to improve the imaginative reasoning quality of the world model. For instance, STORM \cite{zhang2024storm} proposed a Transformer-based \cite{vaswani2017attention} world model that introduces random noise into the Transformer's prediction of environmental states to enhance the robustness of agent training in imagination reasoning. However, it predicts unknown information directly from known states during imagination reasoning, leading to the inability to leverage higher-value differential information for high-quality reasoning. On the other hand, some studies have shown that using long sequence trajectories as input can enhance the imaginative capabilities of the world model, such as in the cases of IRIS \cite{micheli2022transformers} and TWM \cite{robine2023transformer}. However, predicting based on entire long sequences requires processing a large amount of low value information, making it difficult to efficiently infer in long sequence trajectories. Although there have been significant advances in these methods, they have not been able to efficiently leverage the state variation in known trajectories to further improve the imaginative quality of the world model.

To address the challenges outlined above, we explore a novel world model inference framework. Inspired by the human cognitive tendency to focus on variation between pieces of information during reasoning, we propose that enhancing the quality of reasoning-based imaginative interaction in the world model requires equipping them with a similar capacity to infer key information from variation. This potential can be characterized in the world model through the following two capabilities: 
\begin{itemize}
    \item \textbf{Local variation awareness}: This refers to the world model's ability to improve its predictions of unknown information (\textit{i.e.}, reward, termination signal) by perceiving variation between states in adjacent short sequences. 
    \item \textbf{Global variation awareness}: The model needs to efficiently summarize patterns of variation in long sequences, without being distracted by low value information. The quality of imagination in world model can be significantly enhanced by leveraging these patterns to predict future variation.
\end{itemize}

In this work, we introduce \underline{G}lobal-\underline{L}ocal variation \underline{A}wareness \underline{M}amba-based world model (GLAM), a novel model-based reinforcement learning (MBRL) framework that enhances sample efficiency in agent training through high-quality imagination reasoning. In the GLAM, we use Mamba \cite{gu2023mamba} as the inference network and design two modules: the \underline{L}ocal variation awareness \underline{Mamba} module (LMamba) and the \underline{G}lobal variation awareness \underline{Mamba} module (GMamba) to capture variation information. In the inference process GLAM, as illustrated in Fig. \ref{fig: small}, LMamba infers unknown information by predicting variation between adjacent states. Meanwhile, GMamba captures patterns of variation across environmental states from long-sequence trajectories and utilizes these patterns to predict future variation distributions. Specifically, GLAM combines LMamba and GMamba to jointly infer future states and unknown information, thereby achieving variation awareness imagination reasoning. To the best of our knowledge, our work is among the first to emphasize the importance of variation awareness in world model reasoning. Additionally, we design a phased approach to increase the number of interactions between the agent and the world model in each imagined training to enhance the training efficiency of the agent within the world model.

Specifically, our contributions are as follows:

\begin{itemize}
    \item We introduce LMamba, a Mamba-based inference module that enhances the efficiency of world model reasoning for unknown information by perceiving variation.
    \item To improve the accuracy of imagined trajectories, we design GMamba, which enhances the world model's ability to predict variation in future states by capturing the global patterns of variation within known trajectories.
    \item To achieve global-local variation awareness, we design GLAM, a world model framework first employs parallel modules in inference. This framework provides a novel insight into the development of effective world model.

\end{itemize}

\begin{figure*}[t]
\centering
\includegraphics[width=506pt,]
{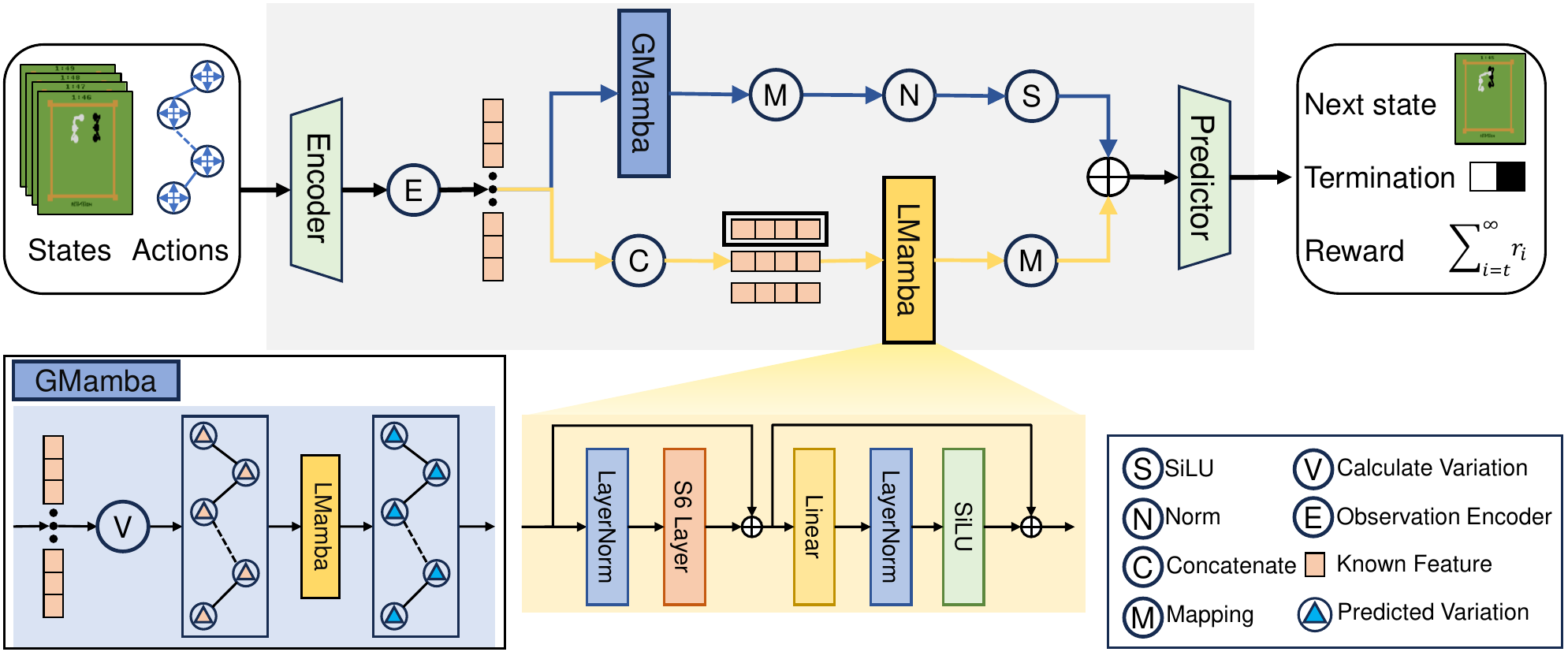}
\vspace{-0.15in}
\caption{Overview of the \underline{G}lobal-\underline{L}ocal variation \underline{A}wareness \underline{M}amba-based world model (GLAM). Based on the known sequence, GLAM infers future information to train the agent in imagination. In each inference, GLAM captures the awareness of global and local variation from long and short sequences and integrates both for prediction.} 
\label{fig: overview}
\end{figure*}

\section{Related Work}

\subsection{World Model}

The \textit{world model} is a type of reinforcement learning method within the realm of Model-Based Reinforcement Learning (MBRL), which is originally proposed in the context of world model \cite{ha2018recurrent} for exploring generative neural network models in RL. 
Under the advantage of training agent in imagination , PlaNet \cite{hafner2019learning} trains agent using fewer training data than model-free RL methods and even outperforms them. Besides, PlaNet introduces the recurrent state space model (RSSM) as a core component. 
Many subsequent works\cite{hafner2019dream,hafner2020mastering,hafner2023mastering} adopt RSSM as inference module. 
Recent research use state space models (SSMs) as the central structure for the world model. S4WM \cite{deng2024facing} builds on S4 \cite{gu2021efficiently} and proposes the first universally applicable world model framework compatible with SSMs. 
Other world model approaches, such as IRIS \cite{micheli2022transformers}, TWM \cite{robine2023transformer}, STORM, and TransDreamer \cite{chen2022transdreamer}, integrate Transformers into the world model, enabling efficient parallel training and leveraging self-attention \cite{vaswani2017attention, dosovitskiy2020image, dong2023heterogeneous} mechanisms for imaginative reasoning. Some works \cite{alonso2024diffusion,ding2024diffusion} utilize the latent diffusion model \cite{rombach2022high, sun2024create} to predict the next state as the world model and achieve advancements in gaming.
World model has also made significant breakshroughs in other areas, such as robotic manipulation \cite{ wu2024pre,seo2022reinforcement, liang2024never}, autonomous driving \cite{10538211,Min_2024_CVPR,Wang_2024_CVPR,pan2023model}, classic control \cite{ha2023dream}.

\subsection{State Space Model}
State space sequence models (SSMs) originated from classical state-space models \cite{kalman1960new}. In recent research, SSMs have proven to be effective in handling long sequences. The latest SSM, Mamba \cite{gu2023mamba}, introduces selective scanning and hardware-aware algorithms to enhance the awareness of sequential data and computational efficiency in the SSM framework. Some works \cite{Yan_2024_CVPR,zhu2024vision,shi2024multi} use SSMs in computer vision tasks, particularly those involving image data. 
For example, \cite{islam2022long,Wang_2023_CVPR,nguyen2022s4nd} achieve breakthrough results by incorporating S4 into various video tasks.

\section{Method}

\subsection{Preliminary}
 State space models (SSMs) originate from a method in continuous systems for handling continuous sequence data. This method accomplishes the mapping of sequences from $x(t) \in \mathbb{R} \Rightarrow y(t) \in \mathbb{R}$ through four weight parameters $\mathbf{A} \in \mathbb{R}^{N\times N}, \mathbf{B} \in \mathbb{R}^{N \times 1}, \mathbf{C} \in \mathbb{R}^{1 \times N}$ and $\mathbf{D} \in \mathbb{R}$, including a hidden state $h(t) \in \mathbb{R}^{N \times 1}$. Here $N$ represents the dimension of parameters. This method can be described by the ordinary differential equation (ODE):
\begin{equation}
    \begin{aligned}
        h(t)&= \mathbf{A}h(t-1) + \mathbf{B}x(t),\\
        y(t)&= \mathbf{C}h(t) + \mathbf{D}x(t),\\
    \end{aligned}
\end{equation}
where $x(t)$ and $y(t)$ represent the continuous input signal and continuous output signal in the time domain.
In order to adapt to computers, SSMs discretize the parameters in the above method. $\mathbf{\Bar{A}, \Bar{B}, \Bar{C}, \Bar{D}}$ represent the corresponding parameters obtained through the zero-order holder, respectively. 

Mamba, also known as S6 layer, removes the parameter matrix $\mathbf{\Bar{D}}$, and its mapping formula is as follows:
\begin{equation}
    \begin{aligned}
        h_t&= \mathbf{\Bar{A}}h_{t-1} + \mathbf{\Bar{B}}x_t,\\
        y_t&= \mathbf{\Bar{C}}h_t.\\
    \end{aligned}
    \label{mamba_cal}
\end{equation}

On the other hand, Mamba introduces a selective scanning algorithm in order to focus on more important segments in the sequence. It replaces $\mathbf{\Bar{B}, \Bar{C}}$ with two higher-dimensional parameter matrices $\mathbf{\Bar{B}, \Bar{C}} \in \mathbb{R}^{N \times N}$ that are mapped by the discretized input $x_t$. The specific formula is as follows: 
\begin{equation}
    \begin{aligned}
        \mathbf{\Bar{B}} &= Linear(x_t), \\
        \mathbf{\Bar{C}} &= Linear(x_t). \\
    \end{aligned}
\end{equation}
This design results in a historical feature vector $h_t \in \mathbb{R}^{N \times N}$ of higher dimension than previous SSMs. For efficient calculation, Mamba introduces the hardware-aware algorithm in inference on sequence. As a result, Mamba is unable to output the historical feature vector $h_t$ in parallel like S5 \cite{smith2022simplified}.

Besides, the selective scanning algorithm enables Mamba to focus more effectively on important segments of the data, compared to the previous SSMs. Although Transformers use a self-attention mechanism for selective scanning, the causal inference form inside Mamba aligns better with the causal relationship where variation triggers environmental information changes in the interaction process. Therefore, we utilize Mamba as the inference network.

We introduce GLAM, a world model that improves the quality of inference through awareness and prediction of variation between states. An overview of GLAM is shown in Fig. \ref{fig: overview}. GLAM leverages two modules, LMamba and GMamba, to perform differential-based reasoning. LMamba infers unknown information and predicts future state by estimating potential variation in the future states. GMamba captures the patterns of variation in environmental state variation over long sequences and uses these patterns to predict the distribution of future variation. Finally, GLAM combines the inference results based on both local and global variation to generate the final prediction of the model.

\subsection{Mamba-based World Model} 
The overall structure of GLAM is shown in Fig. \ref{fig: overview}. The input to the model includes the sequence of observations of the environment $o_{\leq T} = \{o_{t}\}_{t=0}^T$ and the corresponding action sequences $a_{\leq T} = \{a_{t}\}_{t=0}^T$ determined by the agent. Consistent with previous works, we use the image encoder $q_{\phi}$ and the image decoder $p_{\phi}$, which are implemented with convolutional neural networks (CNNs) \cite{lecun1989backpropagation}, to process image observation $o_t$. The feature vector $z_t$ represents the features sampled from $\mathcal{Z}_t$ to represent $o_t$.

\begin{equation}
\begin{aligned}
\label{image_encoder_decoder}
    &\text{Image Encoder:} & z_t&\sim \mathcal{Z}_t = q_{\phi}(o_t),  \\
    &\text{Image Decoder:} & \hat{o}_t&= p_{\phi}(z_t). \\
\end{aligned}
\end{equation}

We design two parallel inference modules, LMamba and GMamba, denoted by $M^l_\phi$ and $M^s_\phi$, to process the global and local variation in sequence, respectively. During the inference process, the known action sequence $a_{\leq T}$, and the image feature sequence $z_{\leq T}=\{z_{t}\}_{t=0}^T$ are mixed by the feature encoder $f_{\phi}$ to obtain the feature vector $e_{\leq T}=\{e_{t}\}_{t=0}^T$. Based on the feature vector, the two modules predict two distributions of the $t$-th moment, $u_t^l$ and $u_t^g$. Here, $u_t^l$ represents the prediction based on the local variation awareness, while $u_t^g$ reflects the prediction based on the variation awareness of the entire known trajectory. Finally, we combine the results to predict future states distribution $\hat{\mathcal{Z}}_{t+1}$ and unknown information $\hat{r_t}$, $\hat{c_t}$, $\hat{o_t}$.
\begin{equation}
\begin{aligned}
\label{world_model_structure}
    &\text{Feature Encoder:}& e_t&= f_\phi(z_{\leq t},a_{\leq t}),  \\
    &\text{GMamba:} & u_t^g&= M^g_\phi(e^g_{\leq t}), \\
    &\text{LMamba:} & u_t^l&= M^l_\phi(e^l_{\leq t}), \\
    &\text{Dynamics Predictor:} & \hat{\mathcal{Z}}_{t+1}&= g^D_\phi(u_t^g, u_t^l), \\
    &\text{Reward Predictor:} & \hat{r}_t&= g^R_\phi(u_t^g, u_t^l), \\
    &\text{Continuation Predictor:} & \hat{c}_t&= g^C_\phi(u_t^g, u_t^l). \\
\end{aligned}
\end{equation}
Here, $f_{\phi}$, $g^D_\phi$, $g^R_\phi$, $g^C_\phi$ are implemented with multi-layer perceptrons (MLPs). 

Inputting a longer known sequence can enhance the quality of imagination in world model. However, the presence of large amounts of redundant information in long sequences consumes significant computational resources and can mislead the model's ability to perceive critical variation. To address this, we clearly define the roles of LMamba and GMamba in processing long sequences. LMamba focuses on inferring unknown information through local discrepancy awareness, and thus, its input is a shorter sequence $e^l_{\leq T} = \{e_{t}\}_{t=T+1-s}^T$ of length $s$. GMamba aims to capture the patterns of global discrepancy changes, and its input is a longer sequence $e^g_{\leq T} = \{e_{t}\}_{t=T+1-l}^T$ of length $l$. In this work, we use fixed length $l=16$ and $s=4$. 

\subsubsection{Local variation awareness Mamba} 
Aiming to enforce the local variation awareness ability of world model, we propose local variation awareness Mamba (LMamba). Unlike other world models that directly inference future state and unknown information, LMamba focuses more on the prediction of variation in feature. It infers future states and unknown information of the current state by predicting the distribution of variation between future states and the current state. This inference method allows LMamba to focus more on improving its awareness of local variation of the environment during training. 

As the input $e^l_{\leq T}$ is a feature sequence that contains both states and actions, LMamba can infer the possible change distribution of the environmental state based on the actions from it, and finally output a sequence, $u^l_{\leq T} = \{u^l_{t}\}_{t=T-s}^T$, of the same length, each vector $u^l_t$ in the sequence serves as the module’s prediction for the $t$-th moment. The inference formula of LMamba is as follows:
\begin{equation}
    \begin{aligned}
        &\text{Mamba:} & s_{\leq t}&= M_\phi(Norm(e^l_{\leq t})),  \\
        & & u^l_{\leq t}&= e_{\leq t} + Norm(s_{\leq t}), \\
    \end{aligned}
    \label{Eq: MWMB}
\end{equation}
where, $Norm$ refers to a normalization method in \cite{ba2016layer} and $M_{\phi}$ represents the Mamba.

Parallel scanning is a key feature for efficiently training the world model \cite{deng2024facing}, and modifications to the dimensions of internal parameter in Mamba pose challenges to this feature in existing SSM-based world model frameworks. Therefore, we design LMamba to directly use the final output of Mamba for prediction during inference, rather than relying on the historical features $h_t$ in Eq. (\ref{mamba_cal}). Our ablation studies demonstrate that even using LMamba as the sole inference module, the world model outperforms the baseline.

\subsubsection{Global variation awareness Mamba} 
To efficiently summarize patterns of variation in long sequences without being distracted by low value information, we developed GMamba, a module designed to be globally aware of variation changes within historical state sequences.

The single-step inference process of GMamba is illustrated in Eq. (\ref{Eq: GMamba}). Based on the input feature sequence $e^g_{\leq T} = \{e_{t}\}_{T-l}^T$, GMamba computes the sequence of variation between states, denoted as $d^g_{\leq T-1} = \{d_{t}\}_{t=T-l}^{T-1}$. Then it analyzes this variation sequence to summarize the distribution patterns of state variation induced by actions in the environment. Finally, GMamba outputs predictions $u^g_{\leq T} = \{u^g_{t}\}_{T+1-l}^T$ based on the awareness of global variation.
\begin{equation}
    \begin{aligned}
        &\text{Global Variation:} & d^g_{\leq t-1}&= e^g_{\leq t}-e^g_{\leq t-1}, \\
        &\text{Mamba:} & s_{t}&= M_\phi(Norm(d^g_{\leq t-1})),  \\
        & & u^g_t&= d^g_{t-1} + Norm(s_{t}). \\
    \end{aligned}
    \label{Eq: GMamba}
\end{equation}

Actually, the changes in variation distribution are more like subtle adjustments on states feature. Therefore, we map $u_t^g$ and $u_t^l$ in different ways.
After being mapped by MLPs, the prediction result $u_t^g$ from $M^g_\phi$ is normalized by LayerNorm \cite{ba2016layer} and activated by SiLU \cite{elfwing2018sigmoid} to retain only the important variation contained in $u_t^l$. The prediction result $u_t^l$ from $M^l_\phi$ just goes through a simple mapping of MLPs, as we need it to retain more state information. 
Specific algorithmic details are demonstrated in Algorithm \ref{alg:parallel_inference_algorithm}. 

During agent training, GLAM only needs to output the prediction of the next moment, as shown in Fig. \ref{fig: overview}. But in the training process of the world model, to achieve parallel inference for $M^l_\phi$, we concatenate the sequence $\{e_i\}_{i=0}^t$ into a short sequence block $E^s = \{ \{e_{i+1-s}, \cdots, e_{i}\}\}_{i=3}^{t}$ and input $E^s$ to $M^l_\phi$ for parallel inference. We demonstrate the parallel inference steps in GLAM in Algorithm \ref{alg:parallel_inference_algorithm}.

\begin{table*}[t]
\centering

\begin{tabular}{l|r|r|r|r|r|r|r|r|r}
    \toprule
    Game & Random & Human & SimPLe & TWM & IRIS & DreamerV3 & Hieros & Storm & GLAM(Ours) \\
    \midrule
    Alien & 228 & 7128 & 617 & 675 & 420 & 959 & 828 & 984 & \textbf{1065}\\
    Amidar & 6 & 1720 & 74 & 122 & 143 & 139 & 127 & \textbf{205} & \textbf{202}\\ 
    Assault & 222 & 742 & 527 & 683 & 1524 & 706 & \textbf{1764} & 801 & 646\\
    Asterix & 210 & 8503 & \textbf{1128} & \textbf{1116} & 854 & 932 & 899 & 1028 &  504\\
    Bank Heist & 14 & 753 & 34 & 467 & 53 & \textbf{649} & 177 & \textbf{641} & 477\\
    Battle Zone & 2360 & 37188 & 4031 & 5068 & 13074 & 12250 & \textbf{15140} & 13540 & 13190\\ 
    Boxing & 0 & 12 & 8 & \textbf{78} & 70 & \textbf{78} & 65 & \textbf{80} & \textbf{76}\\
    Breakout & 2 & 30 & 16 & 20 & \textbf{84} & 31 & 10 & 16 & 11\\
    Chopper Command & 811 & 7388 & 979 & 1697 & 1565 & 420 & 1475 & \textbf{1888} & 1683\\
    Crazy Climber & 10780 & 35829 & 62584 & 71820 & 59234 & \textbf{97190} & 50857 & 66776 & 71872\\
    Demon Attack & 152 & 1971 & 208 & 350 & \textbf{2034} & 303 & 1480 & 165 & 209\\
    Freeway & 0 & 30 & 17 & 24 & \textbf{31} & 0 & \textbf{31} & 0 & 23\\
    Frostbite & 65 & 4335 & 237 & 1476 & 259 & 909 & \textbf{2901} & 1316 & \textbf{2792}\\
    Gopher & 258 & 2413 & 597 & 1675 & 2236 & 3730 & 1473 & \textbf{8240} & 3149\\
    Hero & 1027 & 30826 & 2657 & 7254 & 7037 & \textbf{11161} & 7890 & \textbf{11044} & 7599\\
    James Bond & 29 & 303 & 101 & 362 & 463 & 445 & \textbf{939} & 509 & 738\\
    Kangaroo & 52 & 3035 & 51 & 1240 & 838 & 4098 & \textbf{6590} & 4208 & 1900\\
    Krull & 1598 & 2666 & 2204 & 6349 & 6616 & 7782 & 8130 & 8413 & \textbf{12649}\\
    Kung Fu Master & 256 & 22736 & 14862 & 24555 & 21760 & 21420 & 18793 & 26182 & \textbf{32680}\\
    Ms Pacman & 307 & 6952 & 1480 & 1588 & 999 & 1327 & 1771 & \textbf{2673} & 2094\\
    Pong & -21 & 15 & 13 & \textbf{19} & 15 & \textbf{18} & 5 & 11 & \textbf{18}\\
    Private Eye & 25 & 69571 & 35 & 87 & 100 & 882 & 1507 & \textbf{7781} & 100\\
    Qbert & 164 & 13455 & 1289 & 3331 & 746 & 3405 & 770 & \textbf{4522} & 2434\\
    Road Runner & 12 & 7845 & 5641 & 9109 & 9615 & 15565 & 16950 & 17564 & \textbf{20030}\\
    Seaquest & 68 & 42055 & 683 & \textbf{774} & 661 & 618 & 560 & 525 & 511\\
    Up N Down & 533 & 11693 & 3350 & \textbf{15982} & 3546 & 7667 & $\mathbf{-}$ & 7985 & 7417\\
    \midrule
    Human Mean & 0\% & 100\% & 33\% & 96\% & 105\% & 112\% & 120\% & 122.3\% & \textbf{130.6\%}\\
    Human Median & 0\% & 100\% & 13\% & 51\% & 29\% & 49\% & 56\% & 58.4\% & \textbf{61.8\%}\\
    \bottomrule

\end{tabular}
\caption{Mean scores of models on the 26 games of Atari 100k benchmark, along with the overall human-normalized scores. As per the convention, model results with score fluctuations within $5\%$ are indicated in bold font.
}
\label{result}
\end{table*}

\begin{table}[h] 
    \centering 
    \begin{tabular}{l|c|c|c|c} 
        \toprule
        Game & Pong&Boxing&Kung.&Battle.\\ \midrule
        Dbl.Transformer&4&68&24030& 7600\\
        Dbl.Mamba&3&49&23790& 6150\\ \midrule
        Ours w/o G\&L & 8 & 63 & 26984 &8320   \\ 
        Ours w/o G & 14  &  71  & 29475& 9570 \\ 
        Ours w/o $\mathcal{L}_{VAR}$ & 11 & 75 & 26210&8600   \\  \midrule
        Ours& \textbf{18}  &  \textbf{76} & \textbf{32680} & \textbf{13190}  \\ \bottomrule
    \end{tabular}
    \caption{Ablation study results on inference module.}

    \label{tab:model ablation} 
\end{table}

\subsubsection{Loss Function}
The world model of GLAM incorporates multiple network structures, including Mamba, CNNs, and MLPs. We jointly optimize all networks within the world model in a self-supervised manner. The total loss function value is calculated by the formula in Eq. (\ref{total_loss}). 

\begin{equation}
\begin{aligned}
\label{total_loss}
    \mathcal{L}(\phi)&= \mathcal{L}_{PRED}(\phi) + \mathcal{L}_{DYN}(\phi) + \mathcal{L}_{REP}(\phi) + \mathcal{L}_{VAR}(\phi). \\
\end{aligned}
\end{equation}

The total loss consists of four components. Following DreamerV3, $\mathcal{L}_{PRED}$ represents the world model’s predictive loss for reward $\hat{r_t}$, observation $\hat{o_t}$ and termination signals $\hat{c_t}$. 
The dynamics loss $\mathcal{L}_{DYN}$ emphasizes the model’s prediction of the feature distribution of future states, while the representation loss $\mathcal{L}_{REP}$ slightly adjusts the encoder to approach the predicted feature distribution. $\mathcal{L}_{DYN}$ and $\mathcal{L}_{REP}$ are expressed as Kullback–Leibler (KL) divergences. 
The formula of these three components is as follows:
\begin{equation}
\begin{aligned}
\label{classic_loss}
    \mathcal{L}_{PRED}(\phi)&= \mathcal{L}_{SYM}(\hat{r_t}, r_t)  \\
    & \quad + ||\hat{o_t} - o_t||_2  \\
    & \quad + [c_tlog(\hat{c_t}) + (1-c_t)log(1 - \hat{c_t})],  \\
    \mathcal{L}_{DYN}(\phi)&= \max\left(1, \mathrm{KL}\left[sg(q_{\phi}(o_t)) || g^D_{\phi}(u_t^g, u_t^l)\right]\right), \\
    \mathcal{L}_{REP}(\phi)&= \max\left(1, \mathrm{KL}\left[q_{\phi}(o_t) || sg(g^D_{\phi}(u_t^g, u_t^l))\right]\right), \\
\end{aligned}
\end{equation}
where $\mathcal{L}_{SYM}$ denotes the symlog two-hot loss, $sg(\cdot)$ denotes the operation of stop-gradients.

We specifically design variation loss function for $M^g_\phi$ to constrain its convergence direction relative to $M^l_\phi$. The optimization objective for $M^g_\phi$ is to predict variation information in the next frame based on the pattern in the global sequence. We quantify this objective by measuring the variation between feature vectors across consecutive frames, noted as $\Delta{o_t}$. This loss function is noted as $\mathcal{L}_{VAR}$ in Eq. (\ref{delta_loss}).
\begin{equation}
\begin{aligned}
\label{delta_loss}
    \Delta{o_t} &= sg(z_{t+1} - z_t),  \\
    \mathcal{L}_{VAR}(\phi)&= \max\left(1, \mathrm{KL}\left[\Delta{o_t} || g^D_{\phi}(u^g_t)\right]\right). \\
\end{aligned}
\end{equation}

\begin{algorithm}[tb]
\caption{Parallel inference in GLAM}
\label{alg:parallel_inference_algorithm}
\textbf{Initialize:}:
The input sequence length $l=t$, the length of the concatenate short sequences $s=4$. \\
\textbf{Input}: Observation sequence $o_{0:t}$, Action sequence $a_{0:t}$.

\textcolor{blue}{\textbf{$\triangleright$ Encode feature sequence $e_{0:l}$}:}

~~~~$z_{0:t} \sim \mathcal{Z}_{0:t} = q_{\phi}(o_{0:t})$;

~~~~$e_{0:t} = f_\phi(z_{0:t},a_{0:t})$;

\textcolor{blue}{\textbf{$\triangleright$ Global variation inference}:}

~~~~$ d_{0:t-1}= e_{1:l}-e_{0:t-1}$;

~~~~$u_{1:t}^g = LMamba(d_{0:t-1})$;

~~~~$u_{1:t}^g = LayerNorm(u_{1:t}^g)$;

~~~~$u_{1:t}^g = SiLU(_{1:t}^g)$;

\textcolor{blue}{\textbf{$\triangleright$ Local variation inference}:}

~~~~Concatenate $e_{0:t}$ into a short sequence block $E^s$;

~~~~$\{e_0 \cdots e_t\} \Rightarrow E^s = \{\{ e_{i+1-s}, \cdots, e_{i}\}\}_{i=3}^{t}$;

~~~~$u_{3:t}^g = LMamba(E^s)$;

\textcolor{blue}{\textbf{$\triangleright$ Predict unknown information and next state}:}

~~~~$\hat{\mathcal{Z}}_{4:t+1} = g^D_\phi(u_{3:t}^g, u_{3:t}^l)$;

~~~~$\hat{r}_{3:t} = g^R_\phi(u_{3:t}^g, u_{3:t}^l)$;

~~~~$\hat{c}_{3:t} = g^C_\phi(u_{3:t}^g, u_{3:t}^l)$; 


\textbf{Return}: The unknown information sequence $\hat{r}_{3:t}, \hat{c}_{3:t}$ and the distribution $\hat{\mathcal{Z}}_{4:t+1}$ of the state sequence.

\end{algorithm}

\subsubsection{Variable Number of Interaction}
In GLAM, agent is trained entirely in the imagined
interaction with world model. We found the number of interactions between the agent and the world model in each training session influences the agent's final training performance. Consequently, we incorporate a variable number of interaction during imagination in GLAM. As the steps of world model training increases, the quality of the world model’s imagination gradually improves. Therefore, we adopt a phased approach to increase the number of imagined training within the world model based on the training step. The specific formula for calculating the variable number of interaction is as follows:

\begin{equation}
    \begin{aligned}
    \label{variable_imagine_step}
    n_{t} = \min[n_0 + Int(t/\Delta{n})*n_i, n_{max}],
    \end{aligned}
\end{equation}
where $n_{t}$ represents the variable number of interaction, $t$ denotes the current training step, $n_f$ represents the change frequency of $n_{t}$, $\Delta{n}$ corresponds to the increment of number at each update and $n_{max}$ represents the maximum number of interaction during imagined training. For other parts of the agent’s training, we refer to the method in STORM.

\begin{figure*}[t]
\centering
\includegraphics[width=500pt,]
{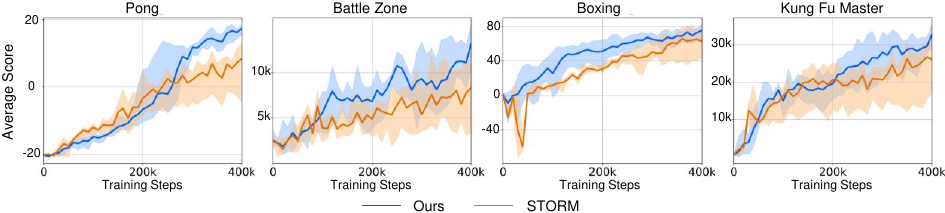}
\vspace{-5pt}
\caption{Comparison of the training results between GLAM and STORM.} 
\label{fig: compare}
\end{figure*}

\section{Experiment}

We evaluate GLAM on a subset of 26 games from Atari 100k \cite{bellemare2013arcade}, a benchmark that is widely used for testing reinforcement learning algorithms. We first introduce the benchmark and the RL methods that are used for comparison, then analyze the comprehensive results, and finally present our ablation study of GLAM.

\subsection{Baselines}
We compare GLAM with the following model-based DRL methods: SimPLe \cite{kaiser2019model}, TWM, IRIS, DreamerV3, Hieros \cite{mattes2023hieros}, and STORM. 
SimPLe and DreamerV3 uses Recurrent State Space Models (RSSM) as the core module for world model inference. Hieros builds a S5-based world model. IRIS, TWM, and STORM uses Transformers as their backbones. IRIS employs long sequences containing all time steps for imagination reasoning, with the sequence length growing as the inference steps progress. TWM performs inference based on fixed-length sequences, while STORM relies only on historical features from the previous time step. 
Similar to previous work, we aim to validate the effectiveness of GLAM, so we do not compare with look-ahead search methods like MuZero \cite{schrittwieser2020mastering} and EfficientZero \cite{ye2021mastering}. 

Same with recent works \cite{zhang2024storm}, we train GLAM using 100k samples in each game. Considering a skip step of 4 frames, the samples correspond to 400k actual game frames, which is about 1.85 hours \cite{zhang2024storm} of real-time game time. The final results of the agent are quantified using the human-normalized score: $S_{Norm}=(S_{Agent}-S_{Random}) / (S_{Human}-S_{Random})$. 

\subsection{Results on Atari 100k}
In our experiment we train GLAM with 5 different seeds in each game.
To evaluate the agent, we perform 20 evaluations for final checkpoints and compute the average scores as the results. The results in Tab. \ref{result} are the average scores obtained using the final checkpoints. 

As shown in Tab. \ref{result}, our method surpasses all baselines in terms of human-normalized scores in mean and median. Notably, GLAM excels in games such as \textit{KungFuMaster, Krull, RoadRunner,} and \textit{Alien}, which are strongly associated with variation. This can be attributed to GMamba and LMamba, enabling the world model to improve reasoning quality through variation awareness. We present a comparison of the training results between GLAM and STORM, the state-of-the-art Transformer-based world model framework, on partial games in Fig. \ref{fig: compare}. 

\begin{figure}
\flushright
\includegraphics[width=235pt,]
{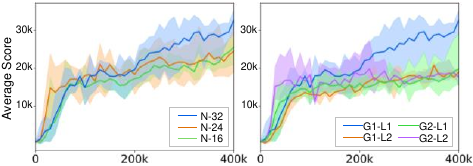}
\caption{Ablation study on variable imagine steps (left) and number of layers in the Mamba (right).} 
\label{fig: ablation3}
\end{figure}

\subsection{Ablation Studies}
\subsubsection{Inference module in GLAM} We conduct ablation studies on the design of the inference modules on \textit{Pong, Boxing, KungFuMaster} and \textit{BattleZone}. \textbf{Dbl.} indicates the combination of two identical modules in direct parallel. To validate the effectiveness of GMamba and LMamba, we choose STORM as the baseline, labeled as \textbf{ours w/o G\&L}. We replace the Transformer in STORM with an equivalent layer number of LMamba, and labeled the new world model as \textbf{ours w/o G}. We also conduct an ablation study on the loss function $\mathcal{L}_{VAR}$ in GMamba. The average scores of the final checkpoints are shown in Tab. \ref{tab:model ablation} and suggest that our design markedly enhances the model's performance.

\begin{figure}
\flushright
\vspace{-0.1in}
\includegraphics[width=235pt]
{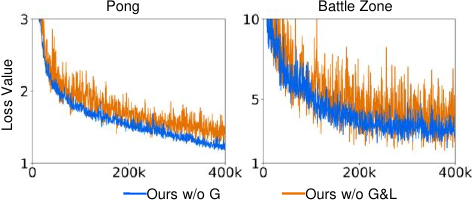}
\vspace{-7pt}
\caption{Loss value in the training process of ours w/o G and ours w/o G\&L.} 
\label{fig: loss}
\end{figure}

In Fig. \ref{fig: loss}, we show the fluctuation of loss values in the training process of \textbf{ours w/o G} and \textbf{ours w/o G\&L}. We observe that the fluctuation amplitude of the convergence curve for the \textbf{ours w/o G} is smaller and more stable than that of \textbf{ours w/o G\&L}. This stability is advantageous for the agent, as it aids in maintaining a consistent direction of convergence during imagined training. We attribute these results to the local awareness capability of LMamba, which minimizes the interference of low value information in inference. 

\subsubsection{Variable Number of Interaction} We explore the impact of different maximum number of interaction on the learning effectiveness of the agent. We design three different parameters of interaction, $n_{max}= 16, 24, 32$, labeled as $N-16,N-24,N-32$ respectively. The variable parameter $n_t$ in training is calculated by Eq. (\ref{variable_imagine_step}), and the parameters $n_0 = 16$ and $\Delta n = 8$ are fixed. Fig. \ref{fig: ablation3} shows the performance of GLAM in \textit{KungFuMaster} under these three conditions. We confirm that during the training process, as the world model continuously converges on environment prediction, adopting a gradually increasing imagination training steps can improve the training efficiency of the agent.

\subsubsection{Number of layers in the Mamba} As an earlier world model framework that utilizes Mamba and parallelly employs two inference models, we explore the impact of different numbers of Mamba layers on training outcomes. We set up four different configurations, labeled $G1-L1, G1-L2, G2-L1$, and $G2-L2$, where $G$ and $L$ represent the number of layers in GMamba and LMamba, respectively. For example, $G1-L1$ indicates using 1 layer of Mamba in both GMamba and LMamba. The training results in \textit{KungFuMaster} are shown in Fig. \ref{fig: ablation3}. Similar to STORM, the performance of model deteriorates in larger-scale model. This decline may be attributed to two factors: 1) The increased complexity of parallel inference modules makes it difficult to achieve synchronized improvement during training, leading to interference in predictions. 2) Minor variation in states can gradually distort as they propagate through multiple layers of the inference module.

\section{Conclusion}
In this study, we introduce GLAM, a world model that incorporates difference awareness in the inference process. GLAM leverages GMamba and LMamba to focus on the global variation patterns and locally variation-related unknowns in the input sequence during inference, respectively. Experimental results demonstrate that GLAM achieves state-of-the-art performance on the Atari 100K benchmark. Furthermore, GLAM is the first world model to implement multi-module parallel inference, offering a novel design paradigm for future world model. 

\subsubsection{Limitations} Mamba provides computational efficiency and effective state representation in capturing long-range dependencies. But our model may still face challenges in handling uncertainty and non-temporality in nonlinear or chaotic systems. Specific limitations include: 1) difficulty in identifying the relationship between state variation and chaotic future change, 2) complex variation patterns that may impede the world model’s ability to produce stable imagine training, 3) agent struggles to explore valuable information. 

\section{Acknowledgments}
We thank all reviewers for their insightful comments and valuable suggestions. This work is supported by the National Natural Science Foundation of China (No.U2013210).

\bibliography{aaai25}

\begin{thebibliography}{45}
\providecommand{\natexlab}[1]{#1}

\bibitem[{Alonso et~al.(2024)Alonso, Jelley, Micheli, Kanervisto, Storkey, Pearce, and Fleuret}]{alonso2024diffusion}
Alonso, E.; Jelley, A.; Micheli, V.; Kanervisto, A.; Storkey, A.; Pearce, T.; and Fleuret, F. 2024.
\newblock Diffusion for World Modeling: Visual Details Matter in Atari.
\newblock \emph{arXiv preprint arXiv:2405.12399}.

\bibitem[{Ba, Kiros, and Hinton(2016)}]{ba2016layer}
Ba, J.~L.; Kiros, J.~R.; and Hinton, G.~E. 2016.
\newblock Layer normalization.
\newblock \emph{arXiv preprint arXiv:1607.06450}.

\bibitem[{Bellemare et~al.(2013)Bellemare, Naddaf, Veness, and Bowling}]{bellemare2013arcade}
Bellemare, M.~G.; Naddaf, Y.; Veness, J.; and Bowling, M. 2013.
\newblock The arcade learning environment: An evaluation platform for general agents.
\newblock \emph{Journal of Artificial Intelligence Research}, 47: 253--279.

\bibitem[{Chen et~al.(2022)Chen, Wu, Yoon, and Ahn}]{chen2022transdreamer}
Chen, C.; Wu, Y.-F.; Yoon, J.; and Ahn, S. 2022.
\newblock Transdreamer: Reinforcement learning with transformer world models.
\newblock \emph{arXiv preprint arXiv:2202.09481}.

\bibitem[{Deng, Park, and Ahn(2024)}]{deng2024facing}
Deng, F.; Park, J.; and Ahn, S. 2024.
\newblock Facing off world model backbones: Rnns, transformers, and S4.
\newblock 36.

\bibitem[{Ding et~al.(2024)Ding, Zhang, Tian, and Zheng}]{ding2024diffusion}
Ding, Z.; Zhang, A.; Tian, Y.; and Zheng, Q. 2024.
\newblock Diffusion world model.
\newblock \emph{arXiv preprint arXiv:2402.03570}.

\bibitem[{Dong et~al.(2023)Dong, Liang, Cong, and Sun}]{dong2023heterogeneous}
Dong, J.; Liang, W.; Cong, Y.; and Sun, G. 2023.
\newblock Heterogeneous forgetting compensation for class-incremental learning.
\newblock In \emph{Proceedings of the IEEE/CVF International Conference on Computer Vision}, 11742--11751.

\bibitem[{Dosovitskiy(2020)}]{dosovitskiy2020image}
Dosovitskiy, A. 2020.
\newblock An image is worth 16x16 words: Transformers for image recognition at scale.
\newblock \emph{arXiv preprint arXiv:2010.11929}.

\bibitem[{Elfwing, Uchibe, and Doya(2018)}]{elfwing2018sigmoid}
Elfwing, S.; Uchibe, E.; and Doya, K. 2018.
\newblock Sigmoid-weighted linear units for neural network function approximation in reinforcement learning.
\newblock \emph{Neural networks}, 107: 3--11.

\bibitem[{Gao et~al.(2024)Gao, Mu, Chen, Duan, Luo, Lu, and Li}]{10538211}
Gao, Z.; Mu, Y.; Chen, C.; Duan, J.; Luo, P.; Lu, Y.; and Li, S.~E. 2024.
\newblock Enhance Sample Efficiency and Robustness of End-to-End Urban Autonomous Driving via Semantic Masked World Model.
\newblock \emph{IEEE Transactions on Intelligent Transportation Systems}, 1--13.

\bibitem[{Gu and Dao(2023)}]{gu2023mamba}
Gu, A.; and Dao, T. 2023.
\newblock Mamba: Linear-time sequence modeling with selective state spaces.
\newblock \emph{arXiv preprint arXiv:2312.00752}.

\bibitem[{Gu, Goel, and R{\'e}(2021)}]{gu2021efficiently}
Gu, A.; Goel, K.; and R{\'e}, C. 2021.
\newblock Efficiently modeling long sequences with structured state spaces.
\newblock \emph{arXiv preprint arXiv:2111.00396}.

\bibitem[{Ha and Schmidhuber(2018)}]{ha2018recurrent}
Ha, D.; and Schmidhuber, J. 2018.
\newblock Recurrent world models facilitate policy evolution.
\newblock 31.

\bibitem[{Ha, Kim, and Kim(2023)}]{ha2023dream}
Ha, J.; Kim, K.; and Kim, Y. 2023.
\newblock Dream to generalize: zero-shot model-based reinforcement learning for unseen visual distractions.
\newblock In \emph{Proceedings of the AAAI Conference on Artificial Intelligence}, volume~37, 7802--7810.

\bibitem[{Hafner et~al.(2019{\natexlab{a}})Hafner, Lillicrap, Ba, and Norouzi}]{hafner2019dream}
Hafner, D.; Lillicrap, T.; Ba, J.; and Norouzi, M. 2019{\natexlab{a}}.
\newblock Dream to control: Learning behaviors by latent imagination.
\newblock \emph{arXiv preprint arXiv:1912.01603}.

\bibitem[{Hafner et~al.(2019{\natexlab{b}})Hafner, Lillicrap, Fischer, Villegas, Ha, Lee, and Davidson}]{hafner2019learning}
Hafner, D.; Lillicrap, T.; Fischer, I.; Villegas, R.; Ha, D.; Lee, H.; and Davidson, J. 2019{\natexlab{b}}.
\newblock Learning latent dynamics for planning from pixels.
\newblock In \emph{International conference on machine learning}, 2555--2565. PMLR.

\bibitem[{Hafner et~al.(2020)Hafner, Lillicrap, Norouzi, and Ba}]{hafner2020mastering}
Hafner, D.; Lillicrap, T.; Norouzi, M.; and Ba, J. 2020.
\newblock Mastering atari with discrete world models.
\newblock \emph{arXiv preprint arXiv:2010.02193}.

\bibitem[{Hafner et~al.(2023)Hafner, Pasukonis, Ba, and Lillicrap}]{hafner2023mastering}
Hafner, D.; Pasukonis, J.; Ba, J.; and Lillicrap, T. 2023.
\newblock Mastering diverse domains through world models.
\newblock \emph{arXiv preprint arXiv:2301.04104}.

\bibitem[{Islam and Bertasius(2022)}]{islam2022long}
Islam, M.~M.; and Bertasius, G. 2022.
\newblock Long movie clip classification with state-space video models.
\newblock In \emph{European Conference on Computer Vision}, 87--104. Springer.

\bibitem[{Kaiser et~al.(2019)Kaiser, Babaeizadeh, Milos, Osinski, Campbell, Czechowski, Erhan, Finn, Kozakowski, Levine et~al.}]{kaiser2019model}
Kaiser, L.; Babaeizadeh, M.; Milos, P.; Osinski, B.; Campbell, R.~H.; Czechowski, K.; Erhan, D.; Finn, C.; Kozakowski, P.; Levine, S.; et~al. 2019.
\newblock Model-based reinforcement learning for atari.
\newblock \emph{arXiv preprint arXiv:1903.00374}.

\bibitem[{Kalman et~al.(1960)}]{kalman1960new}
Kalman, R.~E.; et~al. 1960.
\newblock A new approach to linear filtering and prediction problems [J].
\newblock \emph{Journal of basic Engineering}, 82(1): 35--45.

\bibitem[{LeCun et~al.(1989)LeCun, Boser, Denker, Henderson, Howard, Hubbard, and Jackel}]{lecun1989backpropagation}
LeCun, Y.; Boser, B.; Denker, J.~S.; Henderson, D.; Howard, R.~E.; Hubbard, W.; and Jackel, L.~D. 1989.
\newblock Backpropagation applied to handwritten zip code recognition.
\newblock \emph{Neural computation}, 1(4): 541--551.

\bibitem[{Liang et~al.(2024)Liang, Sun, He, Ren, Dong, and Cong}]{liang2024never}
Liang, W.; Sun, G.; He, Q.; Ren, Y.; Dong, J.; and Cong, Y. 2024.
\newblock Never-Ending Embodied Robot Learning.
\newblock \emph{arXiv preprint arXiv:2403.00336}.

\bibitem[{Mattes, Schlosser, and Herbrich(2023)}]{mattes2023hieros}
Mattes, P.; Schlosser, R.; and Herbrich, R. 2023.
\newblock Hieros: Hierarchical Imagination on Structured State Space Sequence World Models.
\newblock \emph{arXiv preprint arXiv:2310.05167}.

\bibitem[{Micheli, Alonso, and Fleuret(2022)}]{micheli2022transformers}
Micheli, V.; Alonso, E.; and Fleuret, F. 2022.
\newblock Transformers are sample-efficient world models.
\newblock \emph{arXiv preprint arXiv:2209.00588}.

\bibitem[{Min et~al.(2024)Min, Zhao, Xiao, Zhao, Xu, Zhu, Jin, Li, Guo, Xing, Jing, Nie, and Dai}]{Min_2024_CVPR}
Min, C.; Zhao, D.; Xiao, L.; Zhao, J.; Xu, X.; Zhu, Z.; Jin, L.; Li, J.; Guo, Y.; Xing, J.; Jing, L.; Nie, Y.; and Dai, B. 2024.
\newblock DriveWorld: 4D Pre-trained Scene Understanding via World Models for Autonomous Driving.
\newblock In \emph{Proceedings of the IEEE/CVF Conference on Computer Vision and Pattern Recognition (CVPR)}, 15522--15533.

\bibitem[{Nguyen et~al.(2022)Nguyen, Goel, Gu, Downs, Shah, Dao, Baccus, and R{\'e}}]{nguyen2022s4nd}
Nguyen, E.; Goel, K.; Gu, A.; Downs, G.; Shah, P.; Dao, T.; Baccus, S.; and R{\'e}, C. 2022.
\newblock S4nd: Modeling images and videos as multidimensional signals with state spaces.
\newblock 35: 2846--2861.

\bibitem[{Pan et~al.(2023)Pan, Zhu, Zheng, Wang, and Yang}]{pan2023model}
Pan, M.; Zhu, X.; Zheng, Y.; Wang, Y.; and Yang, X. 2023.
\newblock Model-Based Reinforcement Learning with Isolated Imaginations.
\newblock \emph{IEEE Transactions on Pattern Analysis and Machine Intelligence}.

\bibitem[{Robine et~al.(2023)Robine, H{\"o}ftmann, Uelwer, and Harmeling}]{robine2023transformer}
Robine, J.; H{\"o}ftmann, M.; Uelwer, T.; and Harmeling, S. 2023.
\newblock Transformer-based world models are happy with 100k interactions.
\newblock \emph{arXiv preprint arXiv:2303.07109}.

\bibitem[{Rombach et~al.(2022)Rombach, Blattmann, Lorenz, Esser, and Ommer}]{rombach2022high}
Rombach, R.; Blattmann, A.; Lorenz, D.; Esser, P.; and Ommer, B. 2022.
\newblock High-resolution image synthesis with latent diffusion models.
\newblock In \emph{Proceedings of the IEEE/CVF conference on computer vision and pattern recognition}, 10684--10695.

\bibitem[{Schrittwieser et~al.(2020)Schrittwieser, Antonoglou, Hubert, Simonyan, Sifre, Schmitt, Guez, Lockhart, Hassabis, Graepel et~al.}]{schrittwieser2020mastering}
Schrittwieser, J.; Antonoglou, I.; Hubert, T.; Simonyan, K.; Sifre, L.; Schmitt, S.; Guez, A.; Lockhart, E.; Hassabis, D.; Graepel, T.; et~al. 2020.
\newblock Mastering atari, go, chess and shogi by planning with a learned model.
\newblock \emph{Nature}, 588(7839): 604--609.

\bibitem[{Seo et~al.(2022)Seo, Lee, James, and Abbeel}]{seo2022reinforcement}
Seo, Y.; Lee, K.; James, S.~L.; and Abbeel, P. 2022.
\newblock Reinforcement learning with action-free pre-training from videos.
\newblock In \emph{International Conference on Machine Learning}, 19561--19579. PMLR.

\bibitem[{Shi, Dong, and Xu(2024)}]{shi2024multi}
Shi, Y.; Dong, M.; and Xu, C. 2024.
\newblock Multi-Scale VMamba: Hierarchy in Hierarchy Visual State Space Model.
\newblock \emph{arXiv preprint arXiv:2405.14174}.

\bibitem[{Smith, Warrington, and Linderman(2022)}]{smith2022simplified}
Smith, J.~T.; Warrington, A.; and Linderman, S.~W. 2022.
\newblock Simplified state space layers for sequence modeling.
\newblock \emph{arXiv preprint arXiv:2208.04933}.

\bibitem[{Sun et~al.(2024)Sun, Liang, Dong, Li, Ding, and Cong}]{sun2024create}
Sun, G.; Liang, W.; Dong, J.; Li, J.; Ding, Z.; and Cong, Y. 2024.
\newblock Create your world: Lifelong text-to-image diffusion.
\newblock \emph{IEEE Transactions on Pattern Analysis and Machine Intelligence}.

\bibitem[{Sutton(1991)}]{sutton1991dyna}
Sutton, R.~S. 1991.
\newblock Dyna, an integrated architecture for learning, planning, and reacting.
\newblock \emph{ACM Sigart Bulletin}, 2(4): 160--163.

\bibitem[{Vaswani et~al.(2017)Vaswani, Shazeer, Parmar, Uszkoreit, Jones, Gomez, Kaiser, and Polosukhin}]{vaswani2017attention}
Vaswani, A.; Shazeer, N.; Parmar, N.; Uszkoreit, J.; Jones, L.; Gomez, A.~N.; Kaiser, {\L}.; and Polosukhin, I. 2017.
\newblock Attention is all you need.
\newblock 30.

\bibitem[{Vinyals et~al.(2019)Vinyals, Babuschkin, Czarnecki, Mathieu, Dudzik, Chung, Choi, Powell, Ewalds, Georgiev et~al.}]{vinyals2019grandmaster}
Vinyals, O.; Babuschkin, I.; Czarnecki, W.~M.; Mathieu, M.; Dudzik, A.; Chung, J.; Choi, D.~H.; Powell, R.; Ewalds, T.; Georgiev, P.; et~al. 2019.
\newblock Grandmaster level in StarCraft II using multi-agent reinforcement learning.
\newblock \emph{Nature}, 575(7782): 350--354.

\bibitem[{Wang et~al.(2023)Wang, Zhu, Wang, Yu, Liu, Omar, and Hamid}]{Wang_2023_CVPR}
Wang, J.; Zhu, W.; Wang, P.; Yu, X.; Liu, L.; Omar, M.; and Hamid, R. 2023.
\newblock Selective Structured State-Spaces for Long-Form Video Understanding.
\newblock In \emph{Proceedings of the IEEE/CVF Conference on Computer Vision and Pattern Recognition (CVPR)}, 6387--6397.

\bibitem[{Wang et~al.(2024)Wang, He, Fan, Li, Chen, and Zhang}]{Wang_2024_CVPR}
Wang, Y.; He, J.; Fan, L.; Li, H.; Chen, Y.; and Zhang, Z. 2024.
\newblock Driving into the Future: Multiview Visual Forecasting and Planning with World Model for Autonomous Driving.
\newblock In \emph{Proceedings of the IEEE/CVF Conference on Computer Vision and Pattern Recognition (CVPR)}, 14749--14759.

\bibitem[{Wu et~al.(2024)Wu, Ma, Deng, and Long}]{wu2024pre}
Wu, J.; Ma, H.; Deng, C.; and Long, M. 2024.
\newblock Pre-training contextualized world models with in-the-wild videos for reinforcement learning.
\newblock 36.

\bibitem[{Yan, Gu, and Rush(2024)}]{Yan_2024_CVPR}
Yan, J.~N.; Gu, J.; and Rush, A.~M. 2024.
\newblock Diffusion Models Without Attention.
\newblock In \emph{Proceedings of the IEEE/CVF Conference on Computer Vision and Pattern Recognition (CVPR)}, 8239--8249.

\bibitem[{Ye et~al.(2021)Ye, Liu, Kurutach, Abbeel, and Gao}]{ye2021mastering}
Ye, W.; Liu, S.; Kurutach, T.; Abbeel, P.; and Gao, Y. 2021.
\newblock Mastering atari games with limited data.
\newblock 34: 25476--25488.

\bibitem[{Zhang et~al.(2024)Zhang, Wang, Sun, Yuan, and Huang}]{zhang2024storm}
Zhang, W.; Wang, G.; Sun, J.; Yuan, Y.; and Huang, G. 2024.
\newblock STORM: Efficient stochastic transformer based world models for reinforcement learning.
\newblock 36.

\bibitem[{Zhu et~al.(2024)Zhu, Liao, Zhang, Wang, Liu, and Wang}]{zhu2024vision}
Zhu, L.; Liao, B.; Zhang, Q.; Wang, X.; Liu, W.; and Wang, X. 2024.
\newblock Vision mamba: Efficient visual representation learning with bidirectional state space model.
\newblock \emph{arXiv preprint arXiv:2401.09417}.

\end{thebibliography}


\section{Appendices }\label{sec: supplemental_material}

\subsection{More ablation studies}\label{more ablations}

\begin{figure*}[htbp]
\centering
\includegraphics[width=500pt,]
{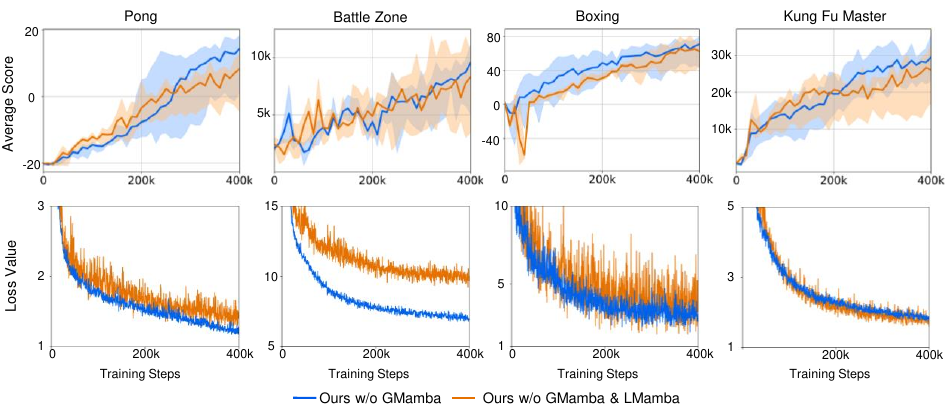}
\caption{The comparison between \textbf{ours w/o G\&L} and \textbf{ours w/o G} on \textit{Pong, Battle Zone, Boxing} and \textit{Kung Fu Master}.} 
\label{fig: ablation-imagine}
\end{figure*}

We present additional ablation experiment results for GMamba and LMamba. The upper part of Fig. \ref{fig: ablation-imagine} shows a comparison of the average training curves for \textbf{ours w/o G\&L} and \textbf{ours w/o G} on \textit{Pong, Battle Zone, Boxing} and \textit{Kung Fu Master}. In the lower part of Fig. \ref{fig: ablation-imagine}, we show the fluctuation of loss values in the training process of two world models in a seed. Furthermore, in the lower part of Fig. \ref{fig: ablation-imagine}, we observed that the fluctuation amplitude of the convergence curve for the \textbf{ours w/o G}B world model is smaller and more stable than that of \textbf{ours w/o G\&L}. This stability is advantageous for the agent, as it aids in maintaining a consistent direction of convergence during imagination training. We attribute these results to the local awareness capability of LMamba, which minimizes the interference of low-value information on inference.

\subsection{Hyperparameters of Mamba World Model}
We present the default hyperparameters of GLAM in our experiment in Table \ref{tab:hyperparameters}. Similar to existing work, in our training of the world model, we assign a weight to each type of loss when calculating the overall loss, as specified in Equation \ref{total_loss2}.
\begin{equation}
\begin{aligned}
\label{total_loss2}
    \mathcal{L}(\phi)&= \mathcal{L}_{PRED}(\phi) + \beta_{1}\mathcal{L}_{DYN}(\phi) \\&+ \beta_{2}\mathcal{L}_{REP}(\phi) + \beta_{3}\mathcal{L}_{VAR}(\phi). \\
\end{aligned}
\end{equation}

\begin{table}[htbp]
    \centering 
    \begin{tabular}{c|c|c} 
        \toprule
        Hyperparameter & Symbol & Value \\ \midrule
        Mamba layers in LMamba  & - &  1  \\ 
        Mamba layers in GMamba & - &  1  \\ 
        Mamba feature dimension & - & 256 \\
        GLAM training batch size & - & 32 \\ 
        GLAM training batch length & - &  64  \\ 
        Weight of $\mathcal{L}_{DYN}(\phi)$ & $\beta_{1}$ & 0.5 \\
        Weight of $\mathcal{L}_{REP}(\phi)$ & $\beta_{2}$ & 0.1 \\
        Weight of $\mathcal{L}_{VAR}(\phi)$ & $\beta_{3}$ & 0.1 \\
        \midrule
        Imagination batch size & - & 1024 \\
        Imagination context length & - & 16 \\
        Imagination initial horizon & $n_0$ & 16 \\
        Horizon increase& $\Delta{n}$ & 8 \\
        Update horizon every env step & $n_i$ & 25000\\
        Maximum horizon & $n_{max}$ & 32 \\ \midrule
        Update GLAM every env step & - & 1\\
        Update agent every env step & - &1\\
        Environment context length & - &16\\
        Optimizer & - &Adam \\
        World model learning rate & - & $1.0\times10^{-4}$ \\
        World model gradient clipping & - &1000 \\
        Actor-critic learning rate & - &$3.0\times10^{-5}$  \\ 
        Actor-critic gradient clipping & - &100  \\
        \bottomrule
    \end{tabular}
    \caption{The default hyperparameters of GLAM.}
    \label{tab:hyperparameters} 
\end{table}

\subsection{Atari video predictions}
\begin{figure*}[h]
\centering
\includegraphics[width=500pt,]
{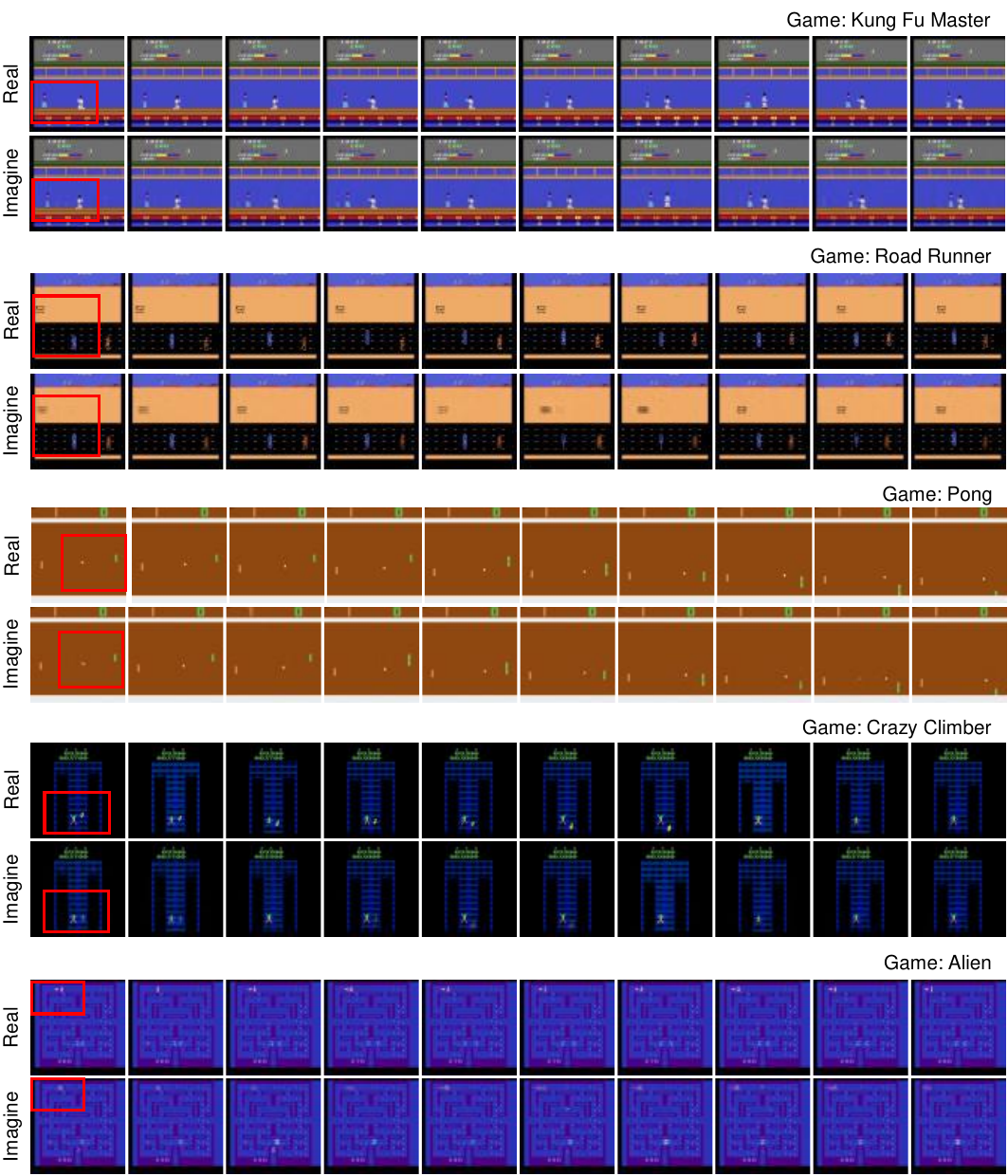}
\caption{The comparison between the imagined trajectories generated by GLAM and the actual game frames.} 
\label{fig: imagine}
\end{figure*}

In Fig. \ref{fig: imagine}, we present the comparison between the imagined trajectories generated by GLAM and the actual game frames across several games. To explore GLAM's ability to predict variations, we focus on localized areas where variations occur within the trajectories (highlighted by red boxes in the figure). We observed that GLAM performs well in predicting variations in objects with distinct features, such as the distances between characters in Kung Fu Master and the highlighted areas in Road Runner and Pong. However, it struggles to accurately predict variations in objects with features that blend into the background, such as the interfering objects in Crazy Climber and the highlighted areas in Alien.

It’s important to note that the detailed accuracy of predicted frames does not fully represent the capability of the world model to train agents. For instance, GLAM achieves state-of-the-art (SOTA) results in both Kung Fu Master and Alien, even though its imagined trajectories in Alien are not entirely accurate. This observation suggests that, in addition to the ability to predict future states, the capability of a world model to predict unknown information (such as rewards and termination signals) from variation data is equally crucial for agent training. This further underscores the importance of LMamba in GLAM, which plays a key role in predicting unknown information through the perception of variations.

\subsection{Computational resources and implementation details} 
In our experiment, we construct our setup using the STORM training framework and the Mamba inference network. We employed a single NVIDIA RTX 4090 GPU with 24GB of VRAM. Training GLAM on an Atari game for 100k frames took approximately 12 hours.

\end{document}